\documentclass[11pt,a4paper]{article}
\usepackage[hyperref]{acl2019}
\usepackage{times}
\usepackage{latexsym}
\usepackage{multirow}
\usepackage{graphicx}
\usepackage{url}

\aclfinalcopy 

\title{UKARA 1.0 Challenge Track 1:\\Automatic Short-Answer Scoring in Bahasa Indonesia}

\author{Ali Akbar Septiandri \\
  Airy \\
  Jakarta, Indonesia \\
  \texttt{ali.septiandri@airy.com} \\\And
  Yosef Ardhito Winatmoko \\
  Jheronimus Academy of Data Science \\
  's-Hertogenbosch, The Netherlands \\
  \texttt{y.a.winatmoko@uvt.nl} \\}

\date{}

\begin{document}
\maketitle
\begin{abstract}
  We describe our third-place solution to the UKARA 1.0 challenge on automated essay scoring. The task consists of a binary classification problem on two datasets | answers from two different questions. We ended up using two different models for the two datasets. For task A, we applied a random forest algorithm on features extracted using unigram with latent semantic analysis (LSA). On the other hand, for task B, we only used logistic regression on TF-IDF features. Our model results in F1 score of 0.812.
\end{abstract}

\section{Introduction}

Automated essay scoring is the application of computers technologies to assist human grader in evaluating the score of written answers \citep{dikli2006overview}. The first track of UKARA 1.0 is the binary classification version of essay scoring, where participants are expected to develop a model that can distinguish right and wrong answers in free text format. The organizer published the questions, the responses with the labels, and the guideline on how to determine whether an answer is acceptable.

During a period of five weeks, the training set and the development set were available and we could validate our model through the score of the development set in a leaderboard. Subsequently, the test set was released, which consists of roughly four times the size of the development set. We are required to submit predicted labels based on the model that we have developed and the winner was determined by the F1 score of the submitted prediction.

Our final submission to this task consists of feature extraction, such as n-grams and TF-IDF, and classical machine learning algorithms, namely logistic regression and random forest. We did not use deep learning at all in our submission. The details of the dataset and our approach will be discussed in the following sections.

\section{Datasets}

The dataset consists of two questions and the respective responses collected by the organizer of the challenge. All questions and responses are in Indonesian. The first question, from now on will be referred as task A, asked about the consequence of the climate change. Concretely, what are the potential problems faced by a climate refugee when they have to migrate to a new place. The second question, referred as task B, is based on an experiment. There were potential customers, initially wanted to buy clothes, who prefer to donate the money instead when they are presented with videos of the clothes manufacturing worker condition before paying. The respondents were required to give their opinion on why do people decided to change their mind. The statistics of the responses for both tasks are shown in Table \ref{tab:dataset-statistics}.

\begin{table}[htbp]
\begin{tabular}{l|ll}
\hline
                   & \textbf{Task A} & \textbf{Task B} \\ \hline
\#Positive Train   & $191 (71\%)$      & $168 (55\%)$      \\
\#Negative Train   & $77 (29\%)$       & $137 (45\%)$      \\
Avg. \#Char        & $87.23$           & $97.33$           \\
\#Dev              & $215$             & $244$             \\
\#Test             & $855$             & $974$             \\ \hline
\end{tabular}
\caption{\label{tab:dataset-statistics}Summary statistics of the dataset.}
\end{table}

\section{Methodology}

\subsection{Preprocessing}

For the preprocessing steps, we first tokenized and lemmatized the text using bahasa Indonesia tokenizer provided by spaCy \citep{spacy2}. We then extracted the features using bag-of-words or TF-IDF. Since the resulting matrix from this feature extraction method tends to be sparse, and to encode token relations, we applied Latent Semantic Analysis (LSA) using Singular Value Decomposition (SVD) \citep{deerwester1990indexing} on the matrix.

Based on our observation, we noticed that the labels of the provided training set are highly inconsistent. Some responses are clearly labelled incorrectly. For illustration, in task A we found "\textit{untuk pindah ke daerah yang aman}" (to move to a safe place) labelled as 1 (correct) while clearly it does not fit the criteria based on the guideline. The mislabeling was even more prominent in task B: "\textit{karana dengan menyumbang kita bisa membuat produksi pakaian menjadi lebih beretika}" (By donating, we can make clothes production becomes more ethical) is considered wrong while "\textit{agar upaya untuk membuat produksi pakaian menjadi lebih beretika.}" (As an effort to make clothes production becomes more ethical) is approved. To approach this problem, we decided to prepare a separate training set with manually corrected labels based on our own judgment. The correction result is shown in Table \ref{tab:corrected-label}. 

Finally, as the responses contain a lot of typo and slang words, we also experimented with simple typo corrector using python difflib package and Indonesian colloquial dictionary \citep{salsabila2018colloquial}. We tried every possible combination of preprocessing steps and whether to use altered version of the training set with a parameter optimization library described in the following subsection.

\begin{table}[htbp]
\begin{tabular}{l|lll}
\hline
                        & \textbf{\begin{tabular}[c]{@{}l@{}}Original \\ Label\end{tabular}} & \textbf{\begin{tabular}[c]{@{}l@{}}Corrected \\ Label\end{tabular}} & \textbf{Count} \\ \hline
\multirow{2}{*}{Task A} & $0$                                                         & $1$                                                          & $10$    \\
                        & $1$                                                         & $0$                                                          & $4$     \\ \hline
\multirow{2}{*}{Task B} & $0$                                                         & $1$                                                          & $46$    \\
                        & $1$                                                         & $0$                                                          & $13$   \\ \hline
\end{tabular}
\caption{\label{tab:corrected-label}Corrected labels of the training set.}
\end{table}

\subsection{The Winning Approach}

After trying several machine learning algorithms, such as k-Nearest Neighbors, Na\"ive Bayes, logistic regression, and random forest, we found that random forest was the best model for task A. This corroborates what was found by \citet{fernandez2014we} in their comprehensive comparisons among several machine learning algorithms on different datasets. On the other hand, logistic regression with L2 regularization was the best for task B. The machine learning library used in this study is scikit-learn \citep{scikit-learn}. Since the dataset is quite small, we used 5-fold cross validation on the training set to avoid overfitting. For our winning approach, we set the \texttt{n\_estimators} parameter for random forest to 200 and keep the default values for other parameters. We also found that it is the best to keep the default parameter values for the logistic regression model.

\subsection{Alternative Approaches}

In parallel, we also experimented with optimization using hyperopt\footnote{http://hyperopt.github.io/hyperopt/} library, which utilizes sequential model-based optimization \citep{bergstra2011algorithms}. We tried to optimize the hyperparameters, including the preprocessing steps, of four different machine learning algorithms: logistic regression, random forest, gradient boosting tree, and support vector machine. We trained separate models for task A and task B. In addition, we tested a voting-based ensemble model by combining all the optimized model of each algorithm. The evaluation metric used for the optimization, including for the voting-ensemble model, is F1 score. The results of the experiments are presented in the next section along with the discussion.

\section{Results and Discussion}

We found that choosing different preprocessing methods resulted in different performances in the two tasks. Therefore, we varied the use of unigram or TF-IDF, and whether we should apply SVD to the resulting matrix. On the other hand, we found that it is always better to use the lemmatizer built on top of spaCy in this task. Moreover, removing stopwords did not contribute much to the performance on the training set. Table of local CV results can be seen in Table~\ref{tab:crossval_a} and Table~\ref{tab:crossval_b}.

\begin{table*}[htbp]
    \centering
    \begin{tabular}{l|lll}
    \hline
                       &          \textbf{Precision} &             \textbf{Recall} &                 \textbf{F1} \\
    \hline
             1-gram+RF &  $0.845 \pm 0.057$ &  $0.921 \pm 0.037$ &  $0.881 \pm 0.035$ \\
         1-gram+logreg &  $0.857 \pm 0.074$ &  $0.869 \pm 0.067$ &  $0.862 \pm 0.065$ \\
         1-gram+SVD+RF &  $0.794 \pm 0.025$ &  $0.984 \pm 0.014$ &  $0.879 \pm 0.014$ \\
     1-gram+SVD+logreg &  $\mathbf{0.859 \pm 0.069}$ &  $0.874 \pm 0.047$ &  $0.866 \pm 0.053$ \\
             TF-IDF+RF &  $0.847 \pm 0.054$ &  $0.942 \pm 0.039$ &  $\mathbf{0.891 \pm 0.036}$ \\
         TF-IDF+logreg &  $0.748 \pm 0.019$ &  $0.979 \pm 0.034$ &  $0.848 \pm 0.022$ \\
         TF-IDF+SVD+RF &  $0.772 \pm 0.025$ &  $\mathbf{0.990 \pm 0.014}$ &  $0.867 \pm 0.014$ \\
     TF-IDF+SVD+logreg &  $0.751 \pm 0.025$ &  $0.979 \pm 0.034$ &  $0.850 \pm 0.025$ \\
    \hline
    \end{tabular}
    \caption{5-fold cross validation results from task A}
    \label{tab:crossval_a}
\end{table*}

\begin{table*}[htbp]
    \centering
    \begin{tabular}{l|lll}
    \hline
                        &          \textbf{Precision} &             \textbf{Recall} &                 \textbf{F1} \\
    \hline
                1-gram+RF &  $0.731 \pm 0.066$ &  $0.744 \pm 0.054$ &  $0.736 \pm 0.047$ \\
            1-gram+logreg &  $\mathbf{0.735 \pm 0.046}$ &  $0.727 \pm 0.080$ &  $0.730 \pm 0.059$ \\
            1-gram+SVD+RF &  $0.686 \pm 0.035$ &  $0.762 \pm 0.046$ &  $0.721 \pm 0.032$ \\
        1-gram+SVD+logreg &  $0.722 \pm 0.067$ &  $0.726 \pm 0.089$ &  $0.723 \pm 0.074$ \\
                TF-IDF+RF &  $0.709 \pm 0.036$ &  $0.750 \pm 0.049$ &  $0.728 \pm 0.034$ \\
            TF-IDF+logreg &  $0.725 \pm 0.035$ &  $0.810 \pm 0.060$ &  $\mathbf{0.764 \pm 0.035}$ \\
            TF-IDF+SVD+RF &  $0.637 \pm 0.026$ &  $\mathbf{0.834 \pm 0.061}$ &  $0.721 \pm 0.036$ \\
        TF-IDF+SVD+logreg &  $0.705 \pm 0.023$ &  $0.809 \pm 0.063$ &  $0.753 \pm 0.036$ \\
    \hline
    \end{tabular}
    \caption{5-fold cross validation results from task B}
    \label{tab:crossval_b}
\end{table*}

For this challenge, we need to optimize the F1 score. Therefore, it is clear from Table~\ref{tab:crossval_b} that we should use the model from TF-IDF with random forest algorithm for task B. From what we can see ini Table~\ref{tab:crossval_a}, we should also go with the same method for task A. However, we decided to be more pessimistic by looking at the largest F1 score after subtracting 1 standard deviation from the mean F1. Thus, we chose 1-gram + SVD + random forest for task A.

Table~\ref{tab:model-result} shows the performance of our best single models compared with two alternative approaches. First, we used the optimized ensemble model trained on the original training set (Ens+Ori). Second, we also use a similar method but with the label-corrected training set (Ens+Upd). While we can see in Table~\ref{tab:model-result} that the F1 score on task B is higher on the label-corrected training set and got a similar result for the development set, the test score is lower than the best single models. The test set labels were most likely as noisy as the training set, thus making the training score with modified label not representative.

\begin{table*}[htbp]
\centering
\begin{tabular}{l|llll}
\hline
        & \textbf{Train A}  & \textbf{Train B}  & \textbf{Dev}  & \textbf{Test} \\
\hline
Best    & $0.879$             & $0.764$             & $0.810$         & $0.812$ \\
Ens+Ori & $0.885$             & $0.764$             & $0.799$         & $0.801$ \\
Ens+Upd & $0.898$             & $0.831$             & $0.810$         & $0.803$ \\
\hline
\end{tabular}
\caption{\label{tab:model-result}F1 score comparison with the alternative Ensemble model.}
\end{table*}

\begin{figure}[htbp]
    \centering
    \includegraphics[width=\columnwidth]{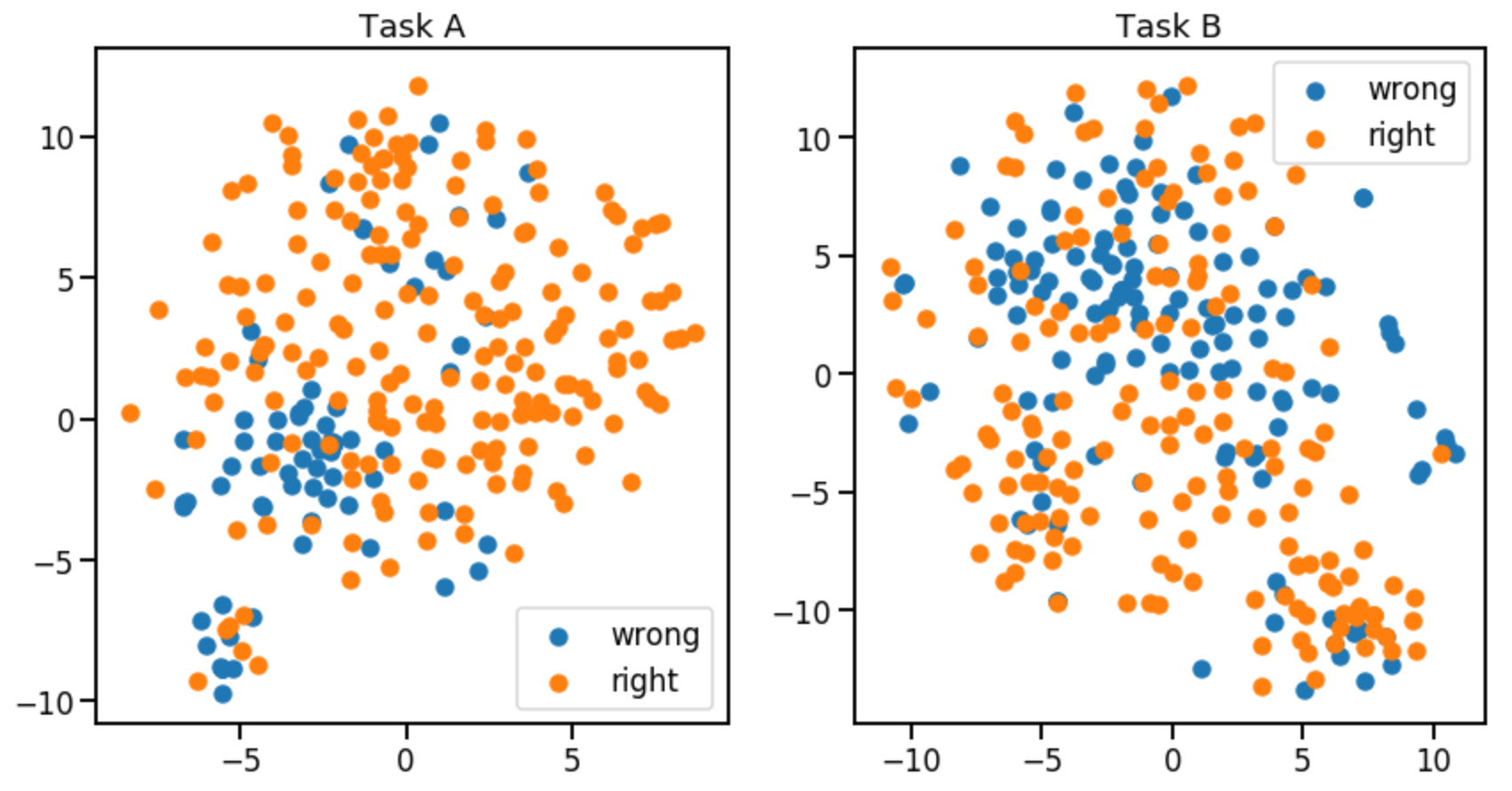}
    \caption{t-SNE visualisation}
    \label{fig:tsne}
\end{figure}

To analyze how hard it is to separate the right from the wrong answers, we reduced the dimensionality of the data into 2D using 1-gram, SVD, and t-SNE \citep{maaten2008visualizing}. Figure~\ref{fig:tsne} suggests that it is harder to separate the two answers in task B. Since most of the answers are short, we also cannot see ``islands'' from applying t-SNE to the data.

\begin{figure}[htbp]
    \centering
    \includegraphics[width=\columnwidth]{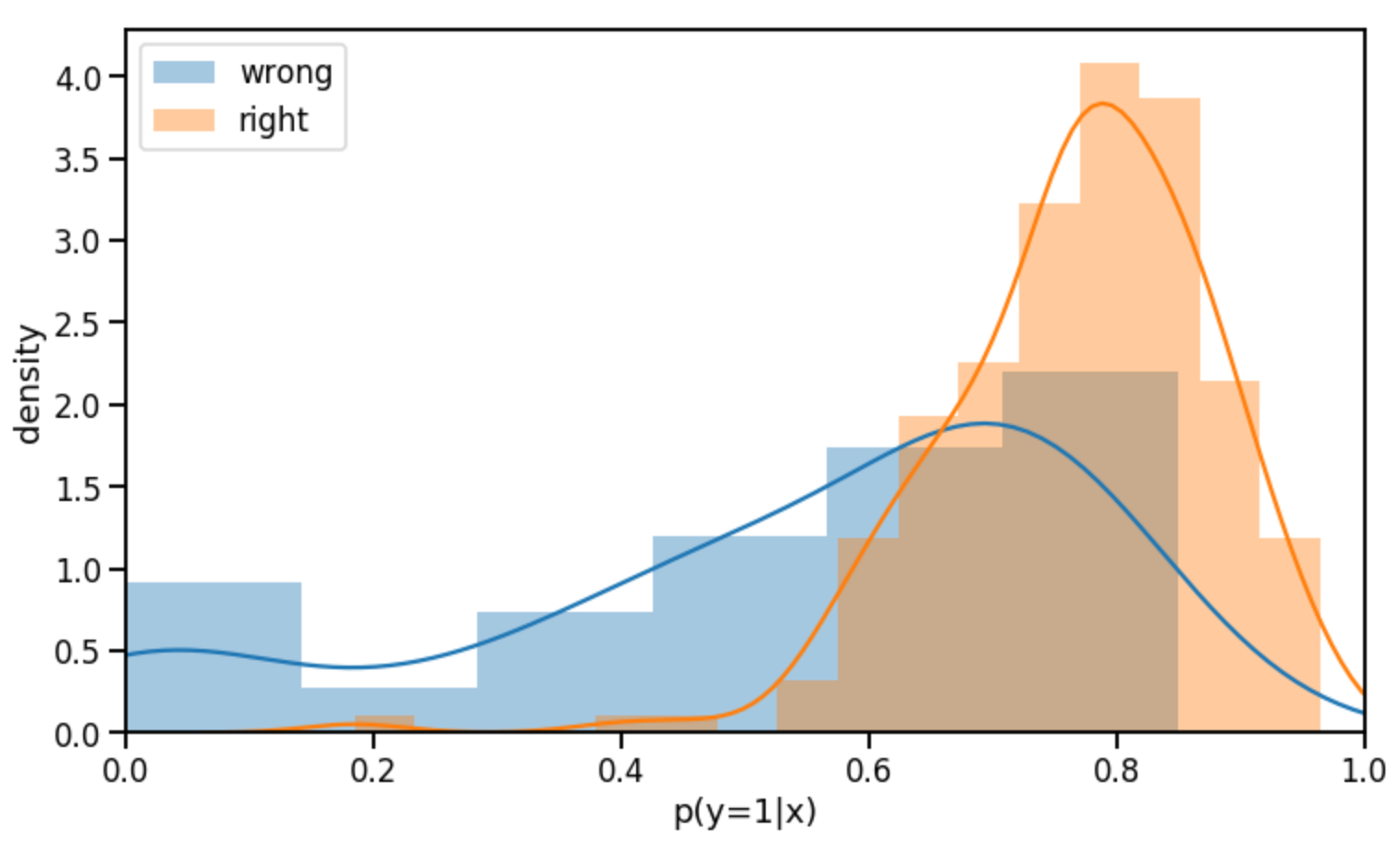}
    \caption{Best model prediction (random forest) with probability on Task A}
    \label{fig:proba-a}
\end{figure}

\begin{figure}[htbp]
    \centering
    \includegraphics[width=\columnwidth]{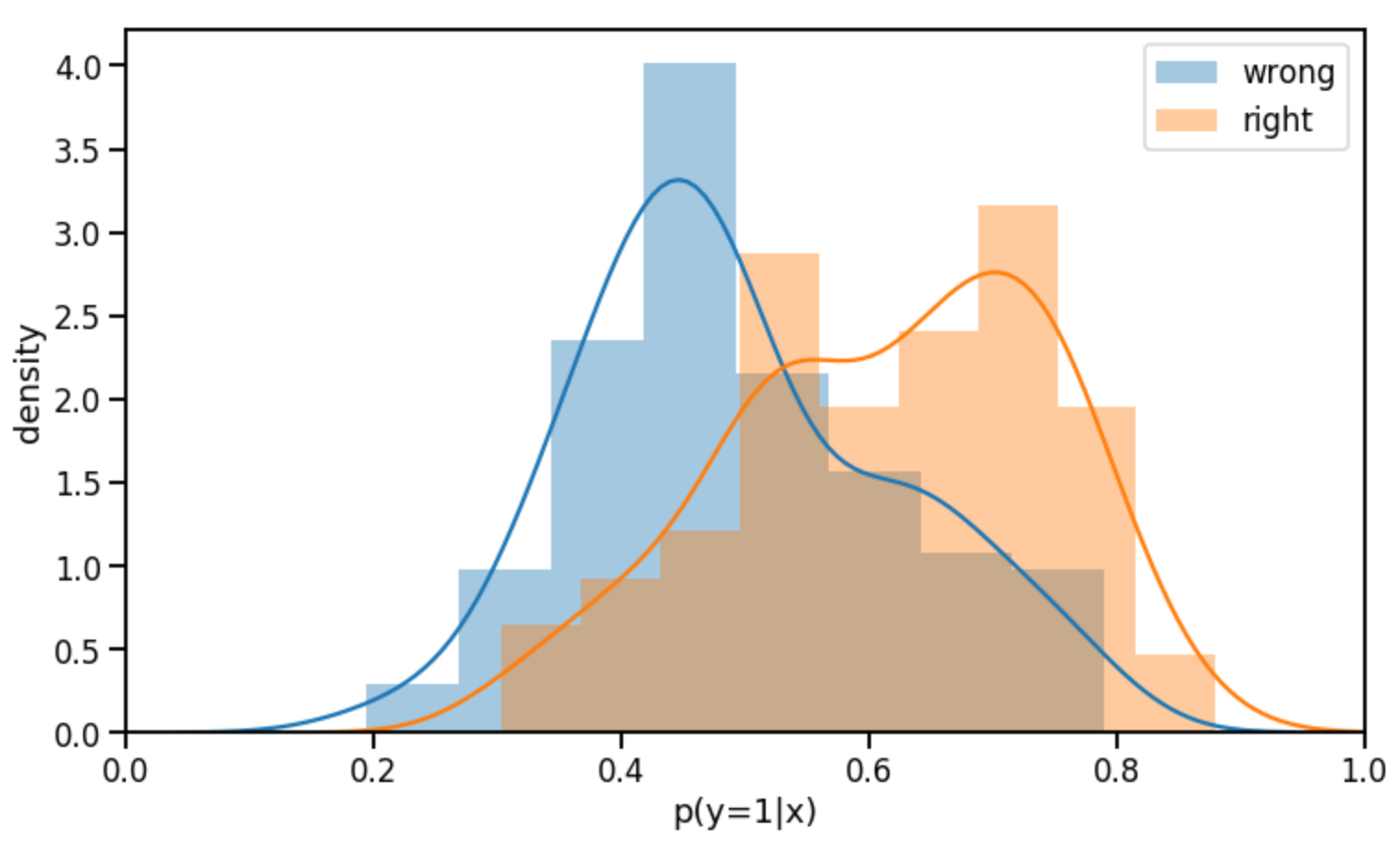}
    \caption{Best model prediction (logistic regression) with probability on Task B}
    \label{fig:proba-b}
\end{figure}

A similar problem was also shown in Figure~\ref{fig:proba-a} and Figure~\ref{fig:proba-b}. We can see a lot more data points in the 0.4-0.6 prediction range in Figure~\ref{fig:proba-b}. This suggests more uncertainty in the model. We argue that this is potentially because of the incorrect labels in the original training set for task B.

\section{Conclusions}
In this report, we describe our winning approach for UKARA 1.0 Challenge Track 1. During the competition, we experimented with single models and ensemble models to predict which answer is correct given an open-ended question. We also tried to re-label the training set and pre-process with text correction which gave a boost in our local cross validation. However, we found that single models with less pre-processing performed better in the test set. The best single model for task A is random forest with unigram+SVD and for task B is logistic regression with TF-IDF. The prediction of both model achieved an overall F1 score of 0.812, which was enough to get us the third position in the final leaderboard.

\bibliography{acl2019}

\begin{thebibliography}{8}
\expandafter\ifx\csname natexlab\endcsname\relax\def\natexlab#1{#1}\fi

\bibitem[{Bergstra et~al.(2011)Bergstra, Bardenet, Bengio, and
  K{\'e}gl}]{bergstra2011algorithms}
James~S Bergstra, R{\'e}mi Bardenet, Yoshua Bengio, and Bal{\'a}zs K{\'e}gl.
  2011.
\newblock Algorithms for hyper-parameter optimization.
\newblock In \emph{Advances in neural information processing systems}, pages
  2546--2554.

\bibitem[{Deerwester et~al.(1990)Deerwester, Dumais, Furnas, Landauer, and
  Harshman}]{deerwester1990indexing}
Scott Deerwester, Susan~T Dumais, George~W Furnas, Thomas~K Landauer, and
  Richard Harshman. 1990.
\newblock Indexing by latent semantic analysis.
\newblock \emph{Journal of the American society for information science},
  41(6):391--407.

\bibitem[{Dikli(2006)}]{dikli2006overview}
Semire Dikli. 2006.
\newblock An overview of automated scoring of essays.
\newblock \emph{The Journal of Technology, Learning and Assessment}, 5(1).

\bibitem[{Fern{\'a}ndez-Delgado et~al.(2014)Fern{\'a}ndez-Delgado, Cernadas,
  Barro, and Amorim}]{fernandez2014we}
Manuel Fern{\'a}ndez-Delgado, Eva Cernadas, Sen{\'e}n Barro, and Dinani Amorim.
  2014.
\newblock Do we need hundreds of classifiers to solve real world classification
  problems?
\newblock \emph{The Journal of Machine Learning Research}, 15(1):3133--3181.

\bibitem[{Honnibal and Montani(2017)}]{spacy2}
Matthew Honnibal and Ines Montani. 2017.
\newblock {spaCy 2}: Natural language understanding with {B}loom embeddings,
  convolutional neural networks and incremental parsing.
\newblock To appear.

\bibitem[{Maaten and Hinton(2008)}]{maaten2008visualizing}
Laurens van~der Maaten and Geoffrey Hinton. 2008.
\newblock Visualizing data using t-sne.
\newblock \emph{Journal of machine learning research}, 9(Nov):2579--2605.

\bibitem[{Pedregosa et~al.(2011)Pedregosa, Varoquaux, Gramfort, Michel,
  Thirion, Grisel, Blondel, Prettenhofer, Weiss, Dubourg, Vanderplas, Passos,
  Cournapeau, Brucher, Perrot, and Duchesnay}]{scikit-learn}
F.~Pedregosa, G.~Varoquaux, A.~Gramfort, V.~Michel, B.~Thirion, O.~Grisel,
  M.~Blondel, P.~Prettenhofer, R.~Weiss, V.~Dubourg, J.~Vanderplas, A.~Passos,
  D.~Cournapeau, M.~Brucher, M.~Perrot, and E.~Duchesnay. 2011.
\newblock Scikit-learn: Machine learning in {P}ython.
\newblock \emph{Journal of Machine Learning Research}, 12:2825--2830.

\bibitem[{Salsabila et~al.(2018)Salsabila, Winatmoko, Septiandri, and
  Jamal}]{salsabila2018colloquial}
Nikmatun~Aliyah Salsabila, Yosef~Ardhito Winatmoko, Ali~Akbar Septiandri, and
  Ade Jamal. 2018.
\newblock Colloquial indonesian lexicon.
\newblock In \emph{2018 International Conference on Asian Language Processing
  (IALP)}, pages 226--229. IEEE.

\end{thebibliography}
\bibliographystyle{acl_natbib}

\appendix


\section{Supplemental Material}
\label{sec:supplemental}

The code for this analysis can be seen on \url{https://github.com/aliakbars/ukara/}.

\end{document}